\pdfoutput=1

\documentclass[11pt]{article}

\usepackage{acl}

\usepackage{times}
\usepackage{latexsym}

\usepackage[T1]{fontenc}

\usepackage[utf8]{inputenc}

\usepackage{microtype}

\usepackage{hyperref}       
\usepackage{url}            
\usepackage{booktabs}       
\usepackage{amsfonts}       
\usepackage{nicefrac}       
\usepackage{xcolor}         

\usepackage{times}
\usepackage{latexsym}
\usepackage{tabularx}
\usepackage[textsize=scriptsize]{todonotes}
\usepackage{multirow}
\usepackage{amsmath}
\usepackage{mathtools}
\usepackage{amssymb}
\usepackage{amsthm}
\usepackage{tablefootnote}
\usepackage{caption}
\usepackage{subcaption}
\usepackage{bbm}
\usepackage{comment}
\usepackage{xspace}
\usepackage{algorithm}
\usepackage{proof}
\usepackage{stmaryrd}
\usepackage{bm}
\usepackage{pifont}
\usepackage{soul}
\usepackage{wrapfig}
\usepackage{scrextend}

\usepackage{algorithm}
\usepackage{bm}
\usepackage[noend]{algpseudocode}

\usepackage{xcolor,colortbl}
\definecolor{tracecolor}{rgb}{0.7216,0.8039,1}
\definecolor{VioletRed}{HTML}{EF58A0}
\definecolor{asli}{HTML}{246FFF}

\newcommand{\STAB}[1]{\begin{tabular}{@{}c@{}}#1\end{tabular}}
\newcommand{\letcat}{ \textsc{LetCat}}
\newcommand{\coin}{ \textsc{CoinFlip}}
\newcommand{\gsm}{ \textsc{GSM}}
\newcommand{\esnli}{ \textsc{e-SNLI}}
\newcommand{\ecqa}{ \textsc{ECQA}}

\DeclareMathOperator*{\argmax}{arg\,max}

\newcommand\ttsmall[1]{\texttt{\small #1}}

\title{Complementary Explanations for Effective In-Context Learning}

\author{Xi Ye$^\diamondsuit$\thanks{\,\, Work done during an internship at Meta AI.} \quad Srinivasan Iyer$^\spadesuit$\quad Asli Celikyilmaz$^\spadesuit$ \quad {Ves Stoyanov}$^\spadesuit$ \\ \textbf{Greg Durrett}$^\diamondsuit$\quad \textbf{Ramakanth Pasunuru}$^\spadesuit$ \\
$^\diamondsuit$ The University of Texas at Austin \quad $^\spadesuit$ Meta AI \\
 $^\diamondsuit${\texttt{\{xiye,gdurrett\}@cs.utexas.edu}} \\
 $^\spadesuit${\texttt{\{sviyer,ves,aslic,rpasunuru\}@meta.com}} \\
}

\begin{document}
\maketitle

\begin{abstract}
Large language models (LLMs) have exhibited remarkable capabilities in learning from explanations in prompts, but there has been limited understanding of exactly how these explanations function or why they are effective.
This work aims to better understand the mechanisms by which explanations are used for in-context learning. We first study the impact of two different factors on the performance of prompts with explanations: the computation trace (the way the solution is decomposed) and the natural language used to express the prompt. By perturbing explanations on three controlled tasks, we show that both factors contribute to the effectiveness of explanations. We further study how to form maximally effective sets of explanations for solving a given test query. We find that LLMs can benefit from the \emph{complementarity} of the explanation set: diverse reasoning skills shown by different exemplars can lead to better performance. Therefore, we propose a maximal marginal relevance-based exemplar selection approach for constructing exemplar sets that are both relevant as well as complementary, which successfully improves the in-context learning performance across three real-world tasks on multiple LLMs.

\end{abstract}

\section{Introduction}

Large language models (LLMs) have achieved promising progress in learning from only a few exemplars in prompts via in-context learning (ICL) \cite{gpt3,palm}. To scale to complex tasks, recent work in the past year has shown that LLMs can benefit from explanations in prompts, particularly for tasks involving multi-step reasoning \cite{scratch,chain,selfconsistency,madaan2022language,jung2022maieutic}.
However, while including explanations in prompts has been demonstrated to be useful,
little has been shown regarding what particular features make them effective and how they function in ICL.

\begin{figure}[t]
\centering
\includegraphics[width=0.95\linewidth,trim=475 120 475 102,clip]{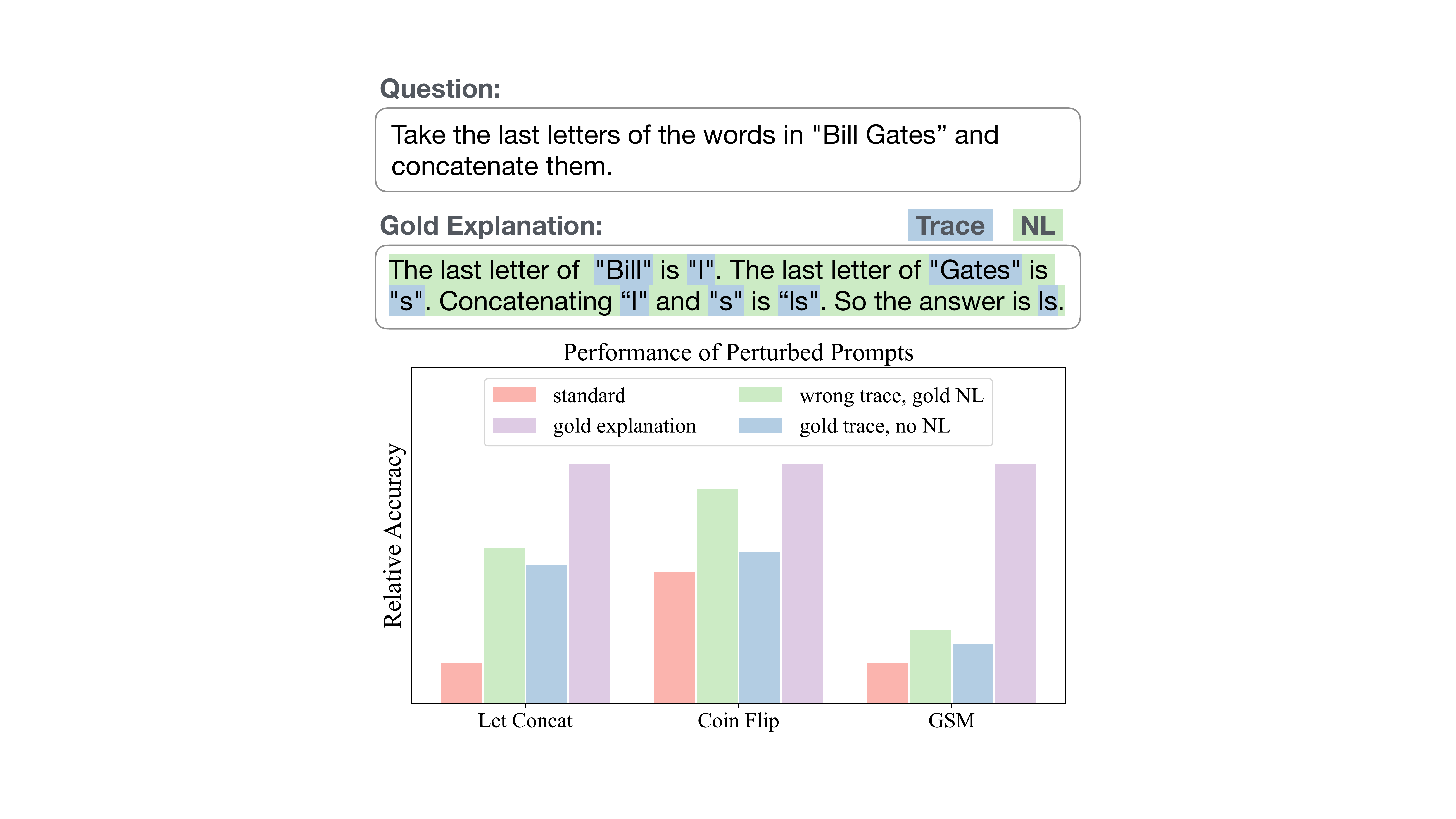}
\caption{Prompting \ttsmall{OPT}~\cite{metaopt} with explanations where we perturb the computation traces or natural language. Perturbing either traces or natural language will lead to performance degradation.}
\label{fig:motivating}
\end{figure}

Our work aims to better understand the mechanisms by which explanations are used for ICL.
As shown in the example for a multi-step reasoning task in Figure~\ref{fig:motivating}, we view an explanation as a combination of a computation trace and natural language which glues together the states in the trace.
We design a series of probing experiments that perturb the explanations (as shown in Figure~\ref{fig:example_perturb}) and test LLMs' performance to understand the sensitivity of LLMs on these two factors.
The results suggest that both factors contribute to making effective explanations, as LLMs see substantial performance degradation when prompted with defective explanations. Nonetheless, incomplete explanations are still beneficial compared to no explanations at all (Figure~\ref{fig:motivating}).
This suggests that LLMs ``faithfully'' follow the reasoning process specified by the explanations to some extent, as opposed to naively following the template patterns while disregarding critical information~\cite{min2022}.

The observations from our probing experiments lead us to focus next on understanding what makes an effective \emph{set} of exemplars with explanations for solving a given test query. 
We primarily focus on two aspects, the exemplar-exemplar interplay (how exemplars work together) and the query-exemplar interplay. On the former, we find that the \emph{complementarity} of the exemplar set is beneficial, as LLMs can fuse different reasoning processes exhibited by individual exemplars in-context. For these studies, we probe LLMs with a mixture of two types of exemplars; each type only specifies a part of the reasoning process (see Figure~\ref{fig:example_gsm_comp} for detailed instances). We also test how the relevance between the query and the exemplars impacts the performance. Choosing the nearest neighbors (NN) of the query to prompt LLMs has been shown to be effective in the standard prompting setting \cite{whatmakes}. Our experiments covering three similarity metrics show that this is also applicable in the setting that prompts LLMs with explanations.

Our analyses inspire us to rethink the exemplar selection process of using explanations for ICL.
The prominent NN-based paradigm only considers relevance \cite{shin2021,whatmakes,rubinlearning}, which could result in selecting mostly similar exemplars. We argue that complementarity should also be considered when constructing explanation-infused prompts. Therefore, we propose an exemplar selection strategy based on the maximum marginal relevance (MMR)~\cite{mmr} approach which selects exemplars that are both relevant as well as \emph{diverse}. The underlying rationale is that a diverse set of exemplars is more likely to showcase complementary reasoning types that are required to illustrate the reasoning required in the query.
We test our MMR-based strategy on three real-world datasets spanning multiple reasoning tasks. On a powerful LLM, \ttsmall{text-davinci-002}, our MMR-based strategy is able to improve the accuracy over the baseline of using random exemplars by 4.0\%, 3.9\%, and 8.5\% on \gsm{}, \ecqa{}, and \esnli{}, respectively.

In summary, our main findings are: (1) We show that both the computation trace and natural language contribute to making effective explanations for ICL.
 (2) We show that LLMs can benefit from exemplar sets that exhibit both complementarity and relevance to a given test query. (3) We propose an MMR-based exemplar selection strategy considering both complementarity and relevance and demonstrate that it is more effective than solely choosing the nearest neighbors.\footnote{Code is available at \url{https://github.com/xiye17/ComplementaryExpl}.}

\section{Background}
\label{sec:background}
\paragraph{In-Context Learning} Our study is focused on the usage of explanations in in-context learning (ICL). Let $q$ be the test query to solve. The standard ICL prompts a language model, $M$, with a set of exemplar input-output pairs, $\{(q_1,a_1)...(q_m,a_m)\}$, and predict an answer $\hat{a}$ for the query:


$$\hat{a}=\argmax_a p_{M}(a \mid q,\{(q_1,a_1)...(q_m,a_m)\}).$$
\normalsize

In addition to just input-output pairs, we can also include explanations (in the style of Scratchpad \cite{scratch} or chain-of-thought \cite{chain}) in prompts, which leads the LLM to generate explanations for its predictions as well:

$$\hat{a}=\argmax_a \sum_{e} p_{M}(a,e \mid q,C),$$
\normalsize

\noindent where $C=\{(q_1,e_1,a_1)...(q_m,e_m,a_m)\}$ is the set of input-explanation-output triplets in prompts.
Ideally, inference in this requires marginalizing out the explanation $e$, which is impractical, especially with LLMs. Following \citet{chain, interpicl}, we employ greedy decoding to make an approximate prediction during inference.\looseness=-1

The end task performance of ICL is sensitive to the selected exemplars \cite{whatmakes}. While much prior work uses a fixed set of manually selected exemplars \cite{chain,selfconsistency}, there is also work devoted to studying how to select more effective exemplars from a pool of exemplars. Given a test query $q$, the task is to select a set of $m$ exemplars from a pool of $n$ exemplars $D=\{(q_1,e_1,a_1)...(q_n,e_n,a_n)\}$ to construct a prompt for solving $q$. We note that this yields varying exemplar sets for different queries.

\paragraph{Datasets \& Large Language Models}
Our analysis is based on model performance on various reasoning datasets.
For probing experiments, we mainly use symbolic reasoning datasets, including 1) \textsc{Letter Concatenation (Let Cat)} \cite{chain} which requires extracting the last letters of two words and then concatenating them, 2) \textsc{Coin Flips} \cite{chain} which reasons about the states of a coin after two steps of operations (flipping or not flipping), and 3) \textsc{Grade School Math (GSM)} \cite{gsm8k} which focuses on grade school-level arithmetic reasoning problems expressed in natural language.

To investigate the effectiveness of different exemplar selection strategies, we use two more textual reasoning datasets, namely \ecqa{} \cite{ecqa} and \esnli{} \cite{esnli}, in addition to \gsm{}. The task of \esnli{} is to decide whether a premise entails a hypothesis. \ecqa{} asks multiple-choice commonsense questions. These three tasks include human-annotated explanations and cover diverse reasoning abilities.

Our experiments cover an array of LLMs, including \ttsmall{OPT-175B} \cite{metaopt}, \ttsmall{GPT-3 (davinci)} \cite{gpt3}, \ttsmall{InstructGPT (text-davinci-001) } \cite{instructgpt}, and \ttsmall{text-davinci-002}. In addition, we also use GPT-3 Codex models \cite{codex} that are finetuned on a large scale of code snippets in our exemplar selection experiments, namely \ttsmall{code-davinci-001} and \ttsmall{code-davinci-002}. Though codex models primarily target code-related applications, we find that they are strong in textual reasoning tasks as well.\looseness=-1 

\section{Do LLMs Follow Explanations?}
\label{sec:probe}

We first investigate what makes explanations effective for LLMs to learn from. We view an explanation as a computation trace ($T$) that is transformed by a natural language function ($L$) which maps the trace to a complete natural language explanation. A computation trace $T$ is the chain of intermediate steps (instantiated as tokens in explanations), $s_1,\ldots,s_n$, that are used to derive the final answer. For instance, the trace for \letcat{} is the two last letters of the two words and the concatenated two-letter tokens; the traces for \gsm{} are the intermediate equations. These computation traces are wrapped by natural language function $L$ to form the final explanation $L(s_1,\ldots,s_n)$, which presumably makes the generation of these traces more ``natural'' with respect to language modeling.

\begin{figure}[t]
    \centering
    \scriptsize
    \renewcommand{\tabcolsep}{0.7mm}

    \begin{tabularx}{\linewidth}{|c|X|}
    \midrule
     \multirow{12}{*}{\STAB{\rotatebox[origin=c]{90}{\letcat{}}}}
         & \textbf{Question:} Take the last letters of the words in "Bill Gates" and concatenate them. \\
         \cmidrule{2-2}
         & \textbf{Gold:} The last letter of \colorbox{tracecolor}{Bill} is \colorbox{tracecolor}{l}. The last letter of \colorbox{tracecolor}{Gates} is \colorbox{tracecolor}{s}. Concatenating \colorbox{tracecolor}{l} and \colorbox{tracecolor}{s} is \colorbox{tracecolor}{ls}. So the answer is ls.\\
         & \textbf{Mask1:} The last letter of Bill is \_. The last letter of Gates is \_. Concatenating l and s is ls. So the answer is ls.\\
         & \textbf{Mask2:} The last letter of Bill is l. The last letter of Gates is s. Concatenating \_ and \_ is \_. So the answer is ls.\\

          & \textbf{Incorrect:} The last letter of Bill is y. The last letter of Gates is e. Concatenating y and e is ye. So the answer is ye.\\
          & \textbf{No NL:} Bill, l. Gates, s. l, s, ls. So the answer is ls.\\
        \midrule

            \multirow{12}{*}{\STAB{\rotatebox[origin=c]{90}{\coin{}}}}
        & \textbf{Question:} A coin is heads up. Ka does not flip the coin. Sal flips the coin. Is the coin still heads up? \\
         \cmidrule{2-2}
         & \textbf{Gold:} The coin started \colorbox{tracecolor}{heads} up. Ka does not flip the coin, so it becomes \colorbox{tracecolor}{heads} up. Sal flips the coin, so it becomes \colorbox{tracecolor}{tails} up. So the answer is no.\looseness=-1 \\
         & \textbf{Mask1:} The coin started heads up. Ka does not flip the coin, so it becomes \_ up. Sal flips the coin, so it becomes tails up. So the answer is no.\\
         & \textbf{Mask2:} The coin started heads up. Ka does not flip the coin, so it becomes heads up. Sal flips the coin, so it becomes \_ up. So the answer is no. \\
         & \textbf{Incorrect:} The coin started heads up. Ka does not flip the coin, so it becomes tails up. Sal flips the coin, so it becomes heads up. So the answer is yes.\\
         & \textbf{No NL:} heads, heads, tails. So the answer is no.\\
        \midrule
    \multirow{17}{*}{\STAB{\rotatebox[origin=c]{90}{\gsm{}}}}
        & \textbf{Question:} Leah had 32 chocolates and her sister had 42. If they ate 35, how many pieces do they have left in total? \\
         \cmidrule{2-2}
         & \textbf{Gold:} Leah had 32 chocolates and Leah's sister had 42. That means there were originally \colorbox{tracecolor}{32+42=74} chocolates. 35 have been eaten. So in total they still have \colorbox{tracecolor}{74-35=39}  chocolates. The answer is 39.\\
         & \textbf{Mask1:} Leah had 32 chocolates and Leah's sister had 42. That means there were originally 32+42=\_ chocolates. 35 have been eaten. So in total they have 
         \_-35=39 chocolates. The answer is 39.\\
         & \textbf{Mask2:} Leah had 32 chocolates and Leah's sister had 42. That means there were originally \_ chocolates. 35 have been eaten. So in total they have 
         \_ chocolates. The answer is 39. \\
         & \textbf{Incorrect:} Leah had 32 chocolates and Leah's sister had 42. That means there were originally 32+42=62 chocolates. 35 have been eaten. So in total they have 
         62-35=27 chocolates. The answer is 27. \\
        & \textbf{No NL:} 32+42=74, 74-35=39. The answer is 39. \\
        \midrule

    \end{tabularx}
    \caption{Examples of gold explanations and perturbed explanations. We perturb the trace in gold explanations (\colorbox{tracecolor}{colored}) by masking intermediate states or substituting them with incorrect values.}
    \label{fig:example_perturb}
\end{figure}

\paragraph{Setup}
We choose the three symbolic reasoning datasets mentioned in Section~\ref{sec:background} for our probing experiments for two reasons. First, LLMs see substantial benefits from including explanations in prompts for these tasks. Second, we can easily manipulate the traces in their explanations. The gold explanations are directly taken from or adapted from \citet{chain}.
More details on the gold explanations used for these tasks can be found in Appendix~\ref{app:goldexpl}.

We experiment on both LLMs trained with vanilla language modeling objectives (\ttsmall{OPT}, \ttsmall{davinci}) and LLMs that are aligned with human expectations via different forms of instruction tuning (\ttsmall{text-davinci-001}, \ttsmall{text-davinci-002}).

\begin{table*}[t]
    \centering
    \scriptsize
    \renewcommand{\tabcolsep}{1.6mm}
    \begin{tabular}{l |cccc | cccc | cccc}
    \toprule
    &\multicolumn{4}{c|}{\letcat{}} &\multicolumn{4}{c|}{\coin{}} &\multicolumn{4}{c}{\gsm{}{}} \\
    & \texttt{OPT} & \texttt{davinci}  & \texttt{txt-01}  & \texttt{txt-02} & \texttt{OPT} & \texttt{davinci}  & \texttt{txt-01}  & \texttt{txt-02} & \texttt{OPT} & \texttt{davinci}  & \texttt{txt-01}  & \texttt{txt-02} \\
    \midrule
    Standard & \phantom{0}8.5 & \phantom{0}8.5 & 10.5 & 16.0 & 51.5 &  83.0  & 68.0 & 99.0 & \phantom{0}5.5 & \phantom{0}7.5 & 11.0 & 26.5 \\
    Gold & 50.0 &  59.0 & 85.0 &100.  & 94.0 & 89.5 & 100. & 100. & 32.5 & 26.0 & 25.0 & 57.5\\
    \cmidrule{1-1}
    Mask1 & 11.0 & 16.0 & 21.5 & 100.  & 71.0 & 88.0& 61.5& 100. & 19.0 & 21.0 & 12.5 & 29.5 \\
    Mask2 & 32.5 & 49.5 & 68.0 & 100.  & 84.0 & 91.5  & 99.0& 100.  & 10.0 & 16.0 &11.5& 27.5\\
    \cmidrule{1-1}
    Random & 10.0 & 25.0& 28.0 & 13.0  & 52.5 & 54.5 & 67.0 & 69.0 
 & \phantom{0}3.0 & \phantom{0}3.0 & \phantom{0}1.0 &  34.5 \\
    Incorrect & 40.0 & 53.0 & 67.5 & 99.5 & 60.5 & 86.0 & 52.0 & 100.& 18.5 & 17.0 & 10.0 & 16.5\\
    \cmidrule{1-1}
    No NL & 29.0 & 15.0 & 46.5 &100.& 59.5 & 86.0 &  99.0 & 100. & \phantom{0}8.0 & 19.5 & 14.5 & 45.5 \\
    \bottomrule
        
    \end{tabular}
    \caption{In-context performance obtained using various perturbed explanations on three datasets. Perturbed explanations achieve inferior performance than complete ones, but many of the perturbed explanations still grant performance gains over standard prompting.}
    \label{tab:perturbstates}
    \vspace{-0.15in}
\end{table*}

\subsection{Explanations or Triggers?}
\label{sec:probetrace}
 We start by investigating whether actual computation traces matter.
If the correctness of the in-context demonstration is unimportant, then that serves as evidence that explanations act as triggers that induce LLMs to follow certain patterns and perform slot-filling.
To study this, we prompt LLMs with perturbed explanations, i.e., by perturbing computational traces and measuring the impact on performance in an ICL setting.

Figure~\ref{fig:example_perturb} shows concrete examples of how we perturb the gold explanations. We experiment with two ways of perturbing the inputs. The first way is masking out the intermediate states by replacing a state $s_i$ (or several states) in $s_1,\ldots,s_n$ with a mask token (e.g., empty string), which tests whether LLMs can implicitly infer the intermediate states. The second way is to replace a state $s_i$ with an incorrect one, which tests whether LLMs can recover the correct computation from corrupted traces.\looseness=-1

\paragraph{Construction of Perturbed Explanations} 

We carefully design the way to mask out the intermediate states. We experiment with various choices of masks in our preliminary experiments. We do not observe large variance caused by different masks, and we choose to use empty string which leads to less performance degradation in general, as our goal here is to probe LLMs' maximum capabilities in recovering the reasoning process from perturbed explanations.
More details on the preliminary experiments on choosing masks can be found in Appendix~\ref{app:choiceofmask}. When constructing incorrect explanations, we also experiment with different sets of random values used to substitute the correct ones.
Furthermore, we also include complete random explanations (taken from other exemplars), which replace the whole gold traces with incorrect ones. 

\paragraph{Results} Table~\ref{tab:perturbstates} shows the results obtained using prompts with various perturbations on the three tasks. First, \emph{LMs are indeed relying on the actual computation traces}. Perturbing the traces of the explanations will lead to performance degradation in various degrees on all these three tasks for \ttsmall{OPT}, \ttsmall{davinci}, and \ttsmall{text-davinci-001}. \ttsmall{text-davinci-002} does not exhibit performance degradation when being prompted with incomplete explanations on the simple tasks, \letcat{} and \coin{}: when the trace is straightforward, a powerful enough LLM is able to ``shortcut'' some particular steps. While for the more challenging tasks, i.e., \gsm{}, \ttsmall{text-002} is also affected by perturbations in explanations. Nonetheless, \emph{LLMs can still benefit from partially complete or partially correct explanations} and outperform standard prompting without using explanations. In particular, on the \letcat{} task, even completely irrelevant random explanations can be beneficial, although they lag gold explanations. Overall, incorrect and incomplete or even totally irrelevant explanations are able to elicit reasoning, but LLMs do rely on gold  explanations to work well.


\subsection{Is Natural Language Necessary?}
\label{sec:probenl}
Next, we question whether the natural language (NL) is really necessary and test whether LLMs can infer the reasoning steps from the computation traces alone. We perturb gold explanations by not wrapping computation traces with natural language transformation $L$, as shown in the examples from Figure~\ref{fig:example_perturb}, and only retain the traces. 

\paragraph{Results} We show the performance obtained by using these prompts in Table~\ref{tab:perturbstates}. \emph{Natural language also plays an essential role in making effective explanations}. Removing the NL leads to substantially worse performance. On \letcat{}, the accuracy of \ttsmall{OPT}, and \ttsmall{davinci} drops by more than 20, respectively, compared to using gold explanations.  On \gsm{}, removing NL consistently leads to performance degradation. Meanwhile, including intermediate states without NL can still improve the performance compared to not using any explanations.\looseness=-1

\subsection{Discussion}
As suggested in the experimental results in Section~\ref{sec:probetrace} and Section~\ref{sec:probenl},
LLMs do generally follow the explanations in the prompts.
Both concrete computation traces and natural language contribute to making effective explanations for ICL. Perturbing certain parts of the explanations will accordingly result in performance degradation, but partial explanations are still beneficial to LMs.


By contrast, recent work shows LLMs are not sensitive to perturbations on the ground-truth input-label mapping in the standard prompting paradigm that does not use explanations \cite{min2022}. Our work shows that LLMs are sensitive to perturbations in the input-explanation mapping and other more subtle perturbations in the explanations. Using explanations in prompts is a promising way to guide LLMs in learning a new task via ICL.

\section{What Makes A Good Exemplar Set?}
Our probing experiments have established how the general factors, computation trace and natural language, impact explanations' effectiveness in ICL. We now study how a set of exemplars, as a whole, functions together in solving a particular test query. We study this problem from two angles, the interplay between exemplars and the interplay between the query and the exemplars.

\subsection{Exemplar-Exemplar Interplay}
\label{sec:probecomp}
As in Section~\ref{sec:probe}, LLMs can learn to follow the reasoning processes as specified in exemplars. As reasoning processes can be composed,
we hypothesize that LLMs might also be able to fuse the reasoning processes of different exemplars together to solve a test query.
We design a set of probing experiments that successfully verify this hypothesis.

\paragraph{Experiment Design} 
At the abstract level, we compare the performance of LLMs when being prompted with three sets of exemplars. The first and second set of exemplars each focuses on a particular part of the reasoning process, and these two parts are disjoint. That is, for a computation trace $s_1,\ldots,s_n$, the first and second set contain exemplars where $s_i$ and $s_j$ are perturbed, and $i\neq j$.
The third set of exemplars includes the mixture from the first and second sets. We test the ICL performance of the prompts constructed from these three types of exemplar sets on the test set that requires combining two types of reasoning. If the third type gives superior performance than the first two types, that means LLMs can pick up the disjoint reasoning and fuse them in-context.

To better illustrate such a hypothesis, we give a concrete example as follows. We have introduced two different types of masked explanations in Figure~\ref{fig:example_perturb} for \letcat{}, where the first type masks the last letter extraction part, and the second type masks the letter concatenation part. These two different masked explanations specify two steps of the reasoning process needed; combining these two steps will yield the complete reasoning steps needed for solving this task. We test whether LLMs can combine these two reasoning steps in-context if being prompted with a mixture of these two corresponding types of masked prompts.\looseness=-1

\begin{figure}[t]
    \centering
    \renewcommand{\tabcolsep}{1.0mm}
    \scriptsize
    \begin{tabularx}{\linewidth}{lX}
    \toprule
        \multirow{4}{*}{\STAB{\rotatebox[origin=c]{90}{Add Only}}}
 & \textbf{Q:} Marion received 20 more turtles than Mia at the animal rescue center. If Mia received 40 turtles, how many turtles did they receive together?\\
         & \textbf{A:} Since Marion received 20 more turtles than Mia, she had 20 + 40 = 60 turtles. The two received 60 + 40 = 100 turtles. The answer is 100.\\
         \midrule
        \multirow{4}{*}{\STAB{\rotatebox[origin=c]{90}{Mul Only}}} & \textbf{Q:} Super Clean Car Wash Company cleans 80 cars per day. They make \$5 per car washed. How much money will they make in 5 days?\\
         & \textbf{A:} Each day they will make 80 * \$5 = \$400. They will make \$400 * 5 = \$2000 in 5 days. The answer is 2000.\\
         \midrule
         \multirow{4}{*}{\STAB{\rotatebox[origin=c]{90}{Add \& Mul}}} & \textbf{Q:} Peter purchased 20 popsicles at \$0.25 each. He also purchased 4 ice cream bars at \$0.50 each. How much did he pay in total in dollars? \\
         & \textbf{A:}  The popsicles cost 0.25 * 20 = 5 dollars. The ice cream bars cost 0.5 * 4 = 2 dollars. He paid 5 + 2 = 7 dollars. The answer is 7. \\
    \bottomrule
    \end{tabularx}
    \caption{Examples of \gsm{} data points that involve only addition operators, only multiplication operators, and both of them at the same time. We note that Add \& Mul are only used at the test time.}
    \label{fig:example_gsm_comp}
\end{figure}

For \gsm{}, we use a more organic way to partition the reasoning process. We separate the reasoning skills needed for a test query based on the operators (addition and multiplication) that are used in the steps. Concretely, we filter the \gsm{} dataset by looking at the provided explanations paired with examples, and obtain disjoint sets that 1) only involves addition operators in the explanation 2) only involves multiplication operators in the explanation (See Figure~\ref{fig:example_gsm_comp} for examples). Next, we test the performance on a test set consisting of examples that require both operators at the same time (Add and Mul in Figure~\ref{fig:example_gsm_comp}). This forms
a test-bed for investigating whether LLMs can better learn to solve problems where both operators are present at the same time while being prompted with the mixture of these two operators, even if no explicit combinations are shown in the prompts.

\paragraph{Setup}
We experiment on the same three datasets as used in Section~\ref{sec:probe}. On \letcat{} and \coin{}, we test whether LLMs can combine the reasoning steps specified in two different types of masked explanations; on \gsm{}, we test whether LLMs can compose addition and multiplication. The mixture type prompts include half of the exemplars from the first type and second type which bear different reasoning. For each setting of \gsm{}, we experiment with 4 different sets of randomly drawn exemplars and report the average. More details about the setting can be found in Appendix~\ref{app:compsetup}.



\begin{table}[t]
    \centering
    \scriptsize
    \renewcommand{\tabcolsep}{1.6mm}
    \begin{tabular}{cl cccc}
    \toprule
    & & \texttt{OPT} & \texttt{davinci}  & \texttt{txt-01}  & \texttt{txt-02}\\
    \midrule
    \multirow{3}{*}{\STAB{\rotatebox[origin=c]{90}{\sc LetCat}}}
    &Mask1 & 11.0 & 16.0& 21.5 &100.\\
    &Mask2 & 32.5 & 49.5& 68.0 &100.\\
    \cmidrule{2-2}
    &Mixture & \bf 37.0 & \bf 56.5 & \bf 82.0 &100.  \\
    \midrule
        \multirow{3}{*}{\STAB{\rotatebox[origin=c]{90}{\sc Coin}}}
    &Mask1 & 71.0 & 88.0 &61.5&100.\\
    &Mask2 & 84.0 & 91.5 &99.0& 100. \\
    \cmidrule{2-2}
    &Mixture  & \bf 93.5 & 91.0&100.&100. \\
    \midrule
    \multirow{3}{*}{\STAB{\rotatebox[origin=c]{90}{\gsm{}}}}
    & AddOnly & \phantom{0}6.8  &13.5& 14.1& 50.3  \\
    & MulOnly & \phantom{0}4.7 & 17.2& 16.7& 50.1 \\
    \cmidrule{2-2}
    & Mixture & \bf \phantom{0}7.0&\bf 18.9 & \bf 18.2 & \bf 52.0  \\
    \bottomrule
    \end{tabular}
    \caption{The accuracy of prompting LLMs with exemplars focusing on single parts of the reasoning or a mixture of them. LLMs achieve better performance when being prompted with exemplars covering multiple aspects of the reasoning process.}
    \label{tab:probecomp}
\end{table}

\paragraph{Results}
As in Table~\ref{tab:probecomp}, on \letcat{}, the prompts with mixed explanations largely surpass Mask1, and outperform Mask2 by 6.0, 7.0, and 12.0 on \ttsmall{OPT}, \ttsmall{davinci}, \ttsmall{text-davinci-001}, respectively. Particularly, the mixture prompts is able to perform roughly on par with the complete prompt (GoldExpl in Table~\ref{tab:perturbstates}) for \letcat{} on \ttsmall{davinci}, and \ttsmall{text-001}.
On \coin{}, using mixture prompts also leads to improvements on \ttsmall{OPT}.

On the realistic \gsm{} dataset, prompting \ttsmall{text-davinci-002} with only addition or multiplication exemplars leads to a performance of 50.3 and 50.1, respectively, whereas prompting with a mixture of these two types of exemplars achieves a better performance of 52.0. On \ttsmall{davinci}, and \ttsmall{text-001}, addition exemplars give worse performance than multiplication exemplars. Nevertheless, including these inferior addition exemplars in prompts together with multiplication exemplars still leads to better performance, as they can complement the reasoning. In general, results on three datasets suggest LLMs are able to fuse the reasoning process that is spread over different exemplars. Therefore, we can expect the exemplars to be able to complement each other and collaborate together to solve the reasoning needed in the test query.\looseness=-1



    

\begin{table*}[t]
    \centering
    \scriptsize
    \begin{tabular}{lcccc|cccc|cccc}
    \toprule
     & \multicolumn{4}{c}{\ttsmall{code-davinci-001}} & \multicolumn{4}{c}{\ttsmall{code-davinci-002} } & \multicolumn{4}{c}{ \ttsmall{text-davinci-002} }\\
       & \gsm{} & \ecqa{} & \esnli{} & \sc Avg & \gsm{} & \ecqa{} & \esnli{} & \sc Avg & \gsm{} & \ecqa{} & \esnli{} & \sc Avg \\

    \midrule
    Random  & 16.3 & 53.6 & 47.2 & 39.0 & 64.6 & 74.7 &  74.9 
& 71.3 &48.8\phantom{$^{*}$} &71.9\phantom{$^{*}$} & 75.1\phantom{$^{*}$} & 65.3\phantom{$^{*}$} \\
\cmidrule{1-1}
        CLS   &16.5& 55.0& 54.1 & 41.8 & 65.4&74.9&74.8 & 71.7 & 50.4\phantom{$^{*}$} & 72.1\phantom{$^{*}$} & 77.4\phantom{$^{*}$} & 66.6\phantom{$^{*}$} \\

    LLM   & \bf 18.5& \bf 56.0 & \bf 57.4 & \bf 43.9 & 65.8 & \bf 76.8 & \bf 81.6 & \bf 74.7 &\bf 52.0$^{*}$ & \bf 74.3$^{*}$ & \bf 83.9$^{*}$ & \bf 70.0\phantom{$^{*}$}\\
    
    BERTScore &\bf 18.5 & 54.6& 53.7 & 42.3 & \bf 66.7 & 75.9& 75.6 & 72.8 & 51.0\phantom{$^{*}$}&72.8\phantom{$^{*}$} & 78.7\phantom{$^{*}$} & 67.6\phantom{$^{*}$} \\
      \bottomrule
    \end{tabular}
    \caption{Comparison between the performance obtained by choosing relevant exemplars using CLS embedding, LM, or BERTScore. {\sc Avg} denotes the average across the three datasets. Selecting relevant exemplars leads to performance improvements, especially when using LLMs themselves to measure the similarity. Using Bertscore also consistently improves the performance across all tasks, even surpassing LM-based scores on \gsm{}. We note that the results on \ttsmall{text-davinci-002} use the LM-based scores provided by \ttsmall{code-davinci-002} (denoted by $^*$).\looseness=-1 }
    \label{tab:shot_selection}
    \vspace{-0.1in}
\end{table*}
\subsection{Query-Exemplar Interplay}
\label{sec:probenn}
Next, we explore how the interplay between the query and exemplars impacts the ICL performance. Recent work has studied how to make good in-context exemplar sets
for a given query in the standard prompting setting: choosing nearest neighbors that are more similar to the query leads to better performance \cite{whatmakes,shin2021}.
Our work investigates how choosing relevant exemplars impact the performance in the setting when using explanations in prompts. We compare the performance obtained by constructing prompts using nearest neighbors against using randomly selected exemplars. The results verify that choosing the nearest neighbors is also beneficial in this setting.

\paragraph{Similarity Measurements} We test three different ways to measure the similarity $\mathcal{S}(q,q_i)$ between a test query $q$ and an exemplar $q_i$.\footnote{The similarity is measured only based on the input part and excludes the explanations part, as we do not have access to the explanation of the query in the test phase.}
\textbf{1) \text{CLS}-based:} \citet{whatmakes} use smaller LMs (e.g., BERT \cite{devlinbert}) to extract the CLS embedding of the input $q$ and $q_i$ and then use cosine similarity to score the embedding pairs, i.e., $\mathrm{cos}(\mathrm{CLS}(q), \mathrm{CLS}(q_i))$.
\textbf{2) LM-based:}  the similarity is given as the probability of generating the query when the language model is conditioned on the exemplar, i.e., $LM(q \mid q_i)$ \cite{shin2021,rubinlearning}.
\textbf{3) BERTScore:} we also experiment with using BERTScore \cite{bertscore} as the similarity score, in addition to the two approaches that are commonly used in prior work.

It is worthwhile to note that measuring similarity using large LLMs is expensive. As it requires querying a large number of query-exemplar pairs.\footnote{For instance, calculating the similarity between 500 queries and a pool of 500 exemplars for a dataset whose typical question token number is 50, would cost \$500 using GPT-3 API (rate: \$0.02/1000 tokens). }

\paragraph{Setup}
We experiment on three realistic datasets, \gsm{}, \ecqa{}, and \esnli{}.
leaving out synthetic tasks which feature formulaic explanations that are all similar to each other.
We set the number of exemplars to be 8 for all three test datasets. We compare the performance of selecting nearest exemplars against that of selecting random exemplars.\looseness=-1

Given the intensive cost of querying LLMs, we set the train exemplars pool size to be 512, and allocate computational resources to experimenting over 4 sets of randomly selected 512 exemplar pools to alleviate the influence of randomness. We focus on more capable LLMs, including \ttsmall{code-davinci-001}, and \ttsmall{code-davinci-002}, and \ttsmall{text-davinci-002}, leaving out \ttsmall{OPT} and \ttsmall{davinci} which have inferior performance. We note that we do not use \ttsmall{text-002} to measure similarity, owing to its high cost. Rather, we take the similarity scores computed by \ttsmall{code-002} and use those for \ttsmall{text-002}. So the performance of \ttsmall{text-002} when using the LM-based measure might be suboptimal given the discrepancy.\looseness=-1

\paragraph{Results}
As shown in Table~\ref{tab:shot_selection}, choosing relevant exemplars is also useful in the setting that includes explanations in prompts. Using the LM-based similarity measurements brings performance improvements across all three datasets, and has the most significant impacts on \esnli{}, though this is achieved with non-negligible computation cost.
Using CLS-embeddings for selecting exemplars mildly improves the performance on \gsm{}  but does not result in any performance gains on \ecqa{}.
The limited improvements can be attributed to the size of the exemplar pools that we use. In our experiments, the size is 512, which is significantly smaller than that in \citet{whatmakes} (typically tens of thousands of exemplars). Nevertheless, this size is large enough for the LM-based method and BERTScore to take advantage of.

In addition, the results suggest that choosing relevant exemplars using BERTScore is also able to improve the performance across all datasets. Specifically, BERTScore-based exemplar selection achieves an accuracy of 66.7 on \gsm{} using \ttsmall{code-002}, which even surpasses the performance of LM-based exemplar selection. While using BERTScore lags the LM-based on \ecqa{} and \esnli{}, it still outperforms choosing random exemplars or CLS-based exemplar selection. Overall, using BERTScore to select the closest exemplars can lead to credible performance improvements while does not require heavy overheads caused by using LLMs to score query-exemplar pairs.

\section{MMR for Exemplar Selection}

\algnewcommand\Input{\textbf{input: }}
\algnewcommand\Output{\textbf{output: }}

\begin{figure}[t]
    \begin{algorithm}[H]
    \footnotesize
    \caption{\footnotesize MMR-Based Exemplar Selection}
    \begin{algorithmic}[1]
    \Procedure{$\textsc{MMRSelect}$}{$D, q, k,\mathcal{S}$}
    \Statex \Input{exemplar pool $D = \{q_1...q_n\}$, test query $q$, number of shots $m$ and similarity measurement $\mathcal{S}$}
    \Statex \Output{selected exemplars $T=\{q_1...q_m\}$}
    \vspace{0.05in}
    \State $\mathbb{S} := [[\mathcal{S}(q_i,q_j)]]_{q_i,q_j \in D}$; \Comment{the pairwise similarity between exemplars in $D$}
    \State $\mathbb{Q} := [\mathcal{S}(q,q_i)]_{q_i \in D}$; \Comment{the similarity between query and exemplars in $T$}
    \State $T := \{\}$; 
    \While{ $|T| < k$}
    \State $\hat{q} := \mathrm{Equation (1)}$;  \Comment{get the next exemplar based on Eq~(\ref{eq:mmreq})}
    \State $T.\texttt{add}(\hat{q})$
    \EndWhile
    \State \Return $T$;
    \EndProcedure
    \end{algorithmic}
    \label{fig:algorithm}
    \end{algorithm}
        \vspace{-0.2in}
\end{figure}


\begin{table*}[t]
    \centering
    \scriptsize
    \begin{tabular}{lcccc|cccc|cccc}
    \toprule
     & \multicolumn{4}{c}{\ttsmall{code-davinci-001}} & \multicolumn{4}{c}{\ttsmall{code-davinci-002} } & \multicolumn{4}{c}{ \ttsmall{text-davinci-002} }\\
      &  \gsm{} & \ecqa{} & \esnli{} & \sc Avg & \gsm{} & \ecqa{} & \esnli{} & \sc Avg & \gsm{} & \ecqa{} & \esnli{} & \sc Avg \\
\midrule
    LLM NN & 18.5& 56.0 & 57.4 & 43.9 & 65.8 & 76.8 & \bf 81.6 & 74.7 & 52.0$^*$ & 74.3$^*$ & \bf 83.9$^*$ & 70.0$^*$\\
    LLM MMR & 18.7 & \bf 57.2 & \bf 59.5 & \bf 45.1 & 67.0 & 77.4 &  81.5 &\bf 75.3 & \bf 52.8$^*$ & \bf 75.3$^*$ & 83.7$^*$ & \bf 70.6$^*$ \\
        \cmidrule{1-1}
    BERTScore NN & 18.5 & 54.6& 53.7 & 42.3 & 66.7 & 75.9& 75.6 & 72.8 &51.0\phantom{$^{*}$} &72.8\phantom{$^{*}$}&  78.7\phantom{$^{*}$} & 67.6\phantom{$^{*}$} \\ 
  BERTScore MMR   &  \bf 19.4 &  56.3 & 53.9 & 43.2 & \bf 68.2 & \bf 78.1&  77.8 &74.7 &52.0\phantom{$^{*}$} & 73.7\phantom{$^{*}$} &78.2\phantom{$^{*}$} & 68.0\phantom{$^{*}$} \\
      \bottomrule
    \end{tabular}
    \caption{Results of using MMR-based exemplar selection strategy on three datasets ({\sc Avg} denotes the average). Using MMR generally selects better exemplars on all datasets, using either LM-based method or BERTScore.  The results on \ttsmall{text-davinci-002} use the LM-based scores provided by \ttsmall{code-davinci-002} (denoted by $^*$).\looseness=-1}
    \label{tab:mmr}
\end{table*}
We have established that emplar-exemplar interplay together with the query-exemplar interplay impacts the performance of using explanations in ICL. This leads us to rethink how to select good exemplars for a given query. Based on our prior analysis on the effects of complementarity and relevance in Section~\ref{sec:probecomp}, we argue that a good set should consist of \emph{relevant} exemplars that collaboratively cover the reasoning skills required for solving the query.

The prominent paradigm, i.e., NN-based exemplar selection strategy, only considers the relevance between the exemplars and the query. Yet, selecting nearest neighbors could result in mostly similar exemplar sets, which can possibly limit collaboration. We argue that complementarity should also be considered in the exemplar selection process, so that the selected set could have a higher chance to illustrate the required reasoning processes.\looseness=-1

In practice, it is tricky to decide whether the reasoning underlying a set of exemplars is complementary categorically. We therefore use diversity as a proxy, since a set of less similar exemplars is arguably more likely to exhibit complementarity. To that end, we propose a maximal-marginal-relevance \cite{mmr} (MMR) based exemplar selection strategy. The idea is to select exemplars that are relevant to the query while being diverse enough to be collaborative. Suppose for the given query $q$, we have already selected a set of exemplars $T=\{q_i\}$, then we will pick up the next exemplar according to:

\small
\begin{equation}
\label{eq:mmreq}
\argmax_{q_j\in D/T} \lambda \mathcal{S} (q, q_j) - (1-\lambda) max_{q_i \in T} \mathcal{S}(q_j, q_i)    
\end{equation}
\normalsize

\noindent where $\mathcal{S}$ denotes similarity and $\lambda$ is a parameter that controls the balance between relevance and diversity. We rely on MMR to iteratively select exemplars from the exemplar pool, as shown in Algorithm~(1). Note that this requires scoring all exemplar pairs within the pool. To run inference over $m$ queries using a pool of $n$ exemplars, MMR requires to score the similarity of $nn+mn$ pairs.



\paragraph{Results} We apply the MMR strategy on top of LM-based method and BERTScore, leaving out the CLS-based approach which has inferior performance. The experimental setup largely follows Section~\ref{sec:probenn}, please refer to Appendix~\ref{app:mmrsetup} for details. 

We show the results in Table~\ref{tab:shot_selection}. In the setting that uses BERTScore, MMR-based selection successfully improves the performance for almost all LLMs for all datasets, compared to using nearest neighbors. On LM-based method, MMR is also able to improve the performance for \gsm{} and \ecqa{} across all LMs, and only marginally under-performs NN for \esnli{}.

In particular, using the MMR-based selection strategy achieves an accuracy of 68.2 and 78.1 on \gsm{} and \ecqa{} respectively, even outperforming LM-based method that requires a large number of queries to the LM. This suggests that BERTScore and MMR as a combination are able to construct effective explanation-infused prompts that approach that of actually querying LLMs. Furthermore, the fact that LLMs achieve better performance from the exemplars selected using our MMR-based method is congruent with our analysis in the previous section: LLMs can exploit complementary explanations.\looseness=-1

\subsection{Analysis}
\paragraph{Impacts of the Trade-off Between Relevance and Diversity}

\begin{table}[t]
    \centering
    \small
    \begin{tabular}{lccc}
        \toprule

    $\lambda$ & \gsm{} & \ecqa{} & \esnli{} \\
    \cmidrule{2-4}
        1.0 &    66.7 &	75.9 &	75.6 \\
        0.8 &  66.9 & 	75.6 &	76.6\\
        0.6 & \bf 68.2	&77.9	& \bf 78.1\\
        0.5 & \bf 68.2	& \bf 78.1	& 77.8 \\
        0.4 & 66.8	&75.7	&76.0 \\
        0.2 & 65.9	& 75.9	& 74.9\\
        0.0 & 63.5 &	75.5	&75.5 \\
        \bottomrule
    \end{tabular}
    \caption{The performance of MMR exemplar selection strategy with varying $\lambda$. }
    \label{tab:my_label}
\end{table}

We conduct an analysis to investigate how the trade-off between diversity and relevance impacts the performance.  We test the performance under varying $\lambda$ on \ttsmall{code-davinci-002} with BERTScore as the similarity metric. We note this is done on one pool of training exemplars. Generally, when $\lambda$ is large (0.8), the performance is similar to NN ($\lambda=1.0$). MMR typically works well with a $\lambda$ of 0.6 or 0.5 (roughly balancing the two terms). The performance starts to degrade while decreasing $\lambda$ from 0.4 to 0, as the selected exemplars are not relevant enough.

\paragraph{Sensitivity to Different Order}

\begin{table}[t]
    \centering
    \small
    \begin{tabular}{lccc}
    \toprule
    & \gsm{} & \ecqa{} & \esnli{} \\
    \cmidrule{2-4}
      Random   & 65.4\textsubscript{1.3} &    74.1\textsubscript{0.5}  &  74.0\textsubscript{1.2}\\
        NN &  68.6\textsubscript{0.7} &   75.4\textsubscript{0.5}  &       75.9\textsubscript{1.1} \\
        MMR & 69.4\textsubscript{1.0}&77.8\textsubscript{0.7}  &  77.8\textsubscript{0.9}\\
        \bottomrule
    \end{tabular}
    \caption{Mean\textsubscript{variance} of the performance across 5 random order. Using better exemplars has more significant impact than varying exemplar order. }
    \label{tab:order}
\end{table}
We have shown choosing exemplar sets using MMR can lead to better ICL performance, which could be affected by other confounders such as the order of exemplars. We conduct experiments to show that using better exemplar sets has more impact than reordering exemplars. Specifically, we experiment with 5 random orders of the exemplar sets for each query and report averaged performance and variance of the accuracy. We note this is done on one pool of training exemplars for each dataset using \ttsmall{code-davinci-002} with BERTScore as the metric. As shown in Table~\ref{tab:order}, MMR is still substantially better than NN and Random under varying order.

\section{Related Work}
The growing scale of pretrained language models has granted them the ability to learn a new task from a few examples via in-context learning \cite{gpt3}.
Various approaches have been proposed to improve ICL in recent years, including meta-tuning LLMs 
\cite{metaicl,chen2022meta}, calibration of ICL \cite{calibrate,Han2022PrototypicalCF}, automatically determining the orders of exemplars \cite{lufantastically}, and alternative formulation of ICL based on PMI \cite{holtzmansurface} or noisy-channel \cite{minnoisy}. More closely related to our work, prior research also contributes to better understanding ICL as Bayesian inference \cite{xie2022an} or experiments that study what makes in-context learning works \cite{min2022}. Our work focuses on understanding the usage of explanations in ICL, as opposed to standard prompting where the LLMs are presented with only input-output pairs.\looseness=-1

In particular, our work is connected to prior research on effective ways for selecting in-context exemplars \cite{shin2021, whatmakes,rubinlearning,Qiu2022EvaluatingTI,su2022selective}. While past work primarily focuses on the effectiveness of using relevant examples in the standard prompting paradigm, we examine the benefits of the complementary exemplars when prompting with explanations. We also propose an MMR-based strategy, which is more effective than the NN-based approach on various LLMs across three tasks.

Lastly, including textual explanations in prompts has exhibited remarkable benefits for LLMs to learn various reasoning tasks. Using Scratchpad \cite{scratch} or Chain-of-Thought \cite{chain} significantly boosts ICL performance on multi-step reasoning tasks such as arithmetical reasoning and symbolic reasoning. Using free-text rationales is also helpful for more unstructured tasks like QA and NLI \cite{interpicl,Wang2022Rationale}. While recent work largely aims to find better ways to prompt LLMs with explanations \cite{zerocot,LeasttoMostPE,Press2022MeasuringAN, Zhou2022TeachingAR}, we focus on analyzing the role of explanations in ICL and what makes effective explanations.


\section{Conclusion}
We have presented a series of studies on what makes effective explanations for in-context learning. We first investigated the impacts of computation traces and natural language in explanations. Through a set of probing experiments, we found that LLMs rely on both of them to effectively learn from explanations. We further examined the interplay among exemplars within prompts and the interplay between exemplars and the query. Our analysis uncovered the benefits of constructing prompts by selecting complementary explanations that are relevant to the query. Lastly, we proposed an MMR-based exemplar selection strategy, which successfully improved the end task performance across three important datasets.

\section{Limitations}

The models chosen in this work are selected to represent the state-of-the-art at the time the work was conducted, and in some cases omit weaker models. For example, our exemplar selection experiments do not cover those LLMs trained with vanilla language models objectives, namely \ttsmall{OPT} and \ttsmall{davinci}, as we find their performance substantially lags \ttsmall{code-davinci-002} and \ttsmall{text-davinci-002}. For the same reason, we only consider the substantially large language models, omitting LLMs of smaller scales (e.g., \ttsmall{text-curie-001}). Running experiments using smaller LMs or vanilla LMs may provide insights into how scale or instruction finetuning impacts the ability of LMs in learning from explanations, but our investigation mainly focus on selecting exemplars to achieve the best in-context learning performance with state-of-the-art models.

In addition, certain aspects of our approach are computationally intensive, particularly using LM-based similarity scores. However, we think this is still feasible in practice: if practitioners are deploying a real-world system, investing more computation upfront to improve its performance is likely in reach for those deploying LLMs in practice.

Finally, our experiments consider a certain subset of NLP reasoning tasks written in English. While we believe the results here transfer to other tasks in this vein which have been frequently used to evaluate LLMs, it is unknown how well they handle other languages, dialects, or genres of text such as social media data.

\section*{Acknowledgments}

Thanks to anonymous reviewers for their helpful feedback and colleagues at Meta AI for helpful discussions. This work was partially supported by NSF CAREER Award IIS-2145280 and the NSF AI Institute for Foundations of Machine Learning (IFML).

\bibliography{custom}

\onecolumn
\newpage
\twocolumn

\appendix

\section{Details of the Explanations Used for \letcat{} and \coin{} on \ttsmall{OPT}}
\label{app:goldexpl}

\begin{figure}[h]
    \centering
    \scriptsize
    \begin{tabularx}{\linewidth}{|c|X|}
    \midrule
     \multirow{7}{*}{\STAB{\rotatebox[origin=c]{90}{\letcat{}}}}
         & \textbf{Question:} Take the last letters of the words in "Bill Gates" and concatenate them. \\
         \cmidrule{2-2}
         & \textbf{Ours:} Add space to "Bill" and get "B i l l", the last letter is l. Add space to "Gates" and get "G a t e s", the last letter is s. Concatenating l and s is ls. So the answer is ls.\\

         & \textbf{Wei et al. (2022):} The last letter of Bill is l. The last letter of Gates is s. Concatenating l and s is ls. So the answer is ls.\\
        \midrule

    \multirow{9}{*}{\STAB{\rotatebox[origin=c]{90}{\coin{}}}}
        & \textbf{Question:} A coin is heads up. Shaunda does not flip the coin. Shalonda flips the coin. Is the coin still heads up? \\
         \cmidrule{2-2}
         & \textbf{Ours:} The coin started heads up. Shaunda does not flip the coin, so it becomes heads up. Shalonda flips the coin, so it becomes tails up. So the answer is no.\\
         & \textbf{Wei et al. (2022):} The coin was flipped by Shalonda. So the coin was flipped 1 time, which is an odd number. The coin started heads up, so after an odd number of flips, it will be tails up. So the answer is no.\\
        \bottomrule
    \end{tabularx}
    \caption{Examples of original chain-of-thoughts from \citet{chain} and ours used for \ttsmall{OPT}. }
    \label{fig:prompteng}
\end{figure}

\begin{table}[h]
    \centering
    \small
    \begin{tabular}{l cc}
    \toprule
      & \letcat{} & \coin{} \\
    \midrule
    Standard & 8.5 & 51.5 \\
    \cmidrule{1-1}
    \cite{chain} & 29.5 & 61.0 \\
    \cmidrule{1-1}
    Ours & 50.0 & 94.0 \\
    \bottomrule
    \end{tabular}
    \caption{Performance of original chain-of-thoughts and our explanations used for \ttsmall{OPT}.}
    \label{tab:detailgold}
\end{table}

For \gsm{}, we directly use the gold explanations provided in \citet{chain}. For \letcat{} and \coin{}, we take the original explanations from \citet{chain} and manually engineered them, as the original ones are sub-optimal for \ttsmall{OPT} and do not lead to credible gains compared to standard prompting.

We show examples of the original explanations (chain-of-thoughts) used in \citet{chain} and the explanation we adapted for \ttsmall{OPT} in Figure~\ref{fig:prompteng}. For \letcat{}, we add another step of tokenizing the two words. For \coin{}, we change the way of decomposing the problem. As shown in Table~\ref{tab:detailgold}, our adapted explanations lead to more substantial performance improvements over standard prompting. We use engineered explanations for the probing experiments on \ttsmall{OPT}, which allows more distinguishable performance differences. We refer readers to the supplementary materials of \citet{chain} for the complete set of exemplars and explanations.

\section{Details of the Choice of Masks}
\label{app:choiceofmask}

\begin{table}[h]
    \centering
    \small
    \begin{tabular}{l ccccc}
    \toprule
     & N/A & [mask] & ? & \_ & Empty Str \\
    \midrule
    Standard & \multicolumn{5}{c}{\phantom{0}8.5\phantom{0}\phantom{0}} \\
    Gold & \multicolumn{5}{c}{59.0\phantom{0}\phantom{0}} \\
    \midrule
    Mask1 & 14.0 & 14.0 & 15.0 & 13.5 & 16.0 \\
    Mask2 & 48.0 & 48.0 & 48.5 & 43.0 & 49.5\\
    \bottomrule
    \end{tabular}
    \caption{Results of using different mask tokens for \letcat{} on \ttsmall{OPT}.}
    \label{tab:maskchoices}
\end{table}

We conduct preliminary experiments on the \letcat{} dataset using \ttsmall{davinci} to determine the choice of masks. We tested masking with "N/A", "[mask]", "?", "\_", and empty string. The results obtained using different masks are shown in Table~\ref{tab:maskchoices}. Whatever masks are used, LLMs see performance degradation compared to gold explanations, but can still learn from partially complete explanations. We use an empty string as the mask token across all datasets, which leads to the least performance degradation.

\section{Details of the Setup for Exemplar-Exemplar Interplay Experiments}
\label{app:compsetup}
For \letcat{}, we experiment with 4 exemplars where the first steps are perturbed, 4 exemplars where the second steps are perturbed and a mixture of 2 from each of these explanations. For \coin{}, we use 8 exemplars and follow the same setting. For the mixture type of prompts, we experiment with 4 random combinations for mixing two types of masked exemplars.

For \gsm{}, we use three types of prompts constructed by 1) 8 addition-only exemplars, 2) 8 multiplication-only exemplars, and 3) a mixture of 4 exemplars from each of the two types. We note that unlike what's in \letcat{} and \coin{} which uses identical exemplars perturbed in different ways, the exemplars in the three sets, for \gsm{} are drawn from different pools. We experiment with 4 different sets of randomly drawn examples and report the average in the setting.
 We note the test set for \gsm{} that requires composing addition and multiplication contains 1,150 data points in total.

\section{Details of the Setup for MMR-Based Exemplar Selection Experiments}
\label{app:mmrsetup}
 We evaluate the effectiveness of our MMR-based exemplar selection strategy on the three realistic datasets used in Section~\ref{sec:probenn}. Also, the experiments on \ttsmall{text-davinci-002} using the LM-based method rely on similarity scores obtained from \ttsmall{code-davinci-002}, given the prohibitive cost needed to run these experiments. We do not tune $\lambda$ in our experiments. $\lambda$ is set to roughly balance the variance among the two terms in Equation~(1), which is 0.5 across all datasets and methods, except for using the LM-based method on \ttsmall{code-davinci-001}. For this particular setting, we set lambda to be $\frac{2}{3}$, as we observe higher variance among the diversity with exemplars.
 
\section{Details of Prompts for Real-world Datasets}
We showcase how we format the prompts for \gsm{}, \ecqa{}, and \esnli{} in Figure~\ref{fig:prompt_example}, with one exemplar for each of the three datasets. We note the prompt format for \esnli{} is taken from PromptSource \cite{bach2022promptsource}.

\begin{figure}[h]
    \centering
    \footnotesize
    \begin{tabularx}{\linewidth}{|X|}
    \toprule
    \multicolumn{1}{|c|}{\gsm{}} \\
    \midrule
 Q: Leah had 32 chocolates and her sister had 42. If they ate 35, how many pieces do they have left in total? \\
         A: Leah had 32 chocolates and Leah's sister had 42. That means there were originally 32 + 42 = 74 chocolates. 35 have been eaten. So in total they still have 74 - 35 = 39 chocolates. The answer is 39.\\
    \midrule
    \multicolumn{1}{|c|}{\ecqa{}}  \\
    \midrule
    Q: Where can you get a bugle to take home with you?\\
    Answer Choices:\\
    (a) farmer's wife\\
    (b) music store\\
    (c) military base\\
    (d) military band\\
    (e) american army. \\
         A: Bugle is a musical instrument. Musical instruments are available in a music store. Music store is a building. So the answer is (b).\\
    \midrule
    \multicolumn{1}{|c|}{\esnli{}}  \\
    \midrule
        Premise:\\
        "A man at a flea market browsing."\\
        Based on this premise, can we conclude the hypothesis "A man is sleeping at a flea market." is true?\\
        OPTIONS:\\
        - yes\\
        - no \\
        - not possible to tell \\
         A: One cannot be sleeping and browsing at the same time. The answer is no.\\
    \bottomrule
    \end{tabularx}
    \caption{Detailed examples of prompts for \gsm{}, \ecqa{}, and \esnli{}.}
    \label{fig:prompt_example}
\end{figure}

\section{License of Datasets}
\begin{itemize}
    \item {\sc GSM} \cite{gsm8k}: MIT license.
    \item {\sc e-SNLI} \cite{esnli}: MIT license.
    \item {\sc ECQA} \cite{ecqa}:Community Data License Agreement - Sharing - Version 1.0.
    \item {\sc Letter Concatenation} and {\sc CoinFlip} \cite{chain}:  MIT license.
\end{itemize}

\end{document}